\documentclass[conference]{IEEEtran}
\IEEEoverridecommandlockouts

\usepackage{cite}
\usepackage{amsmath,amssymb,amsfonts}
\usepackage{algorithmic}
\usepackage{graphicx}
\usepackage{textcomp}
\usepackage{xcolor}
\usepackage{url}
\usepackage{booktabs}
\usepackage{listings}
\usepackage{array}

\newcommand{\FixedHeight}{3.5cm}

\lstdefinestyle{codeinrow}{
	language=Python,
	basicstyle=\ttfamily\scriptsize,
	numbers=none,
	frame=single,
	breaklines=true,
	showstringspaces=false,
	columns=fullflexible,
	keepspaces=true,
	aboveskip=0pt,
	belowskip=0pt
}

\newenvironment{FixedCodeBox}[1][\FixedHeight]{%
	\begin{minipage}[c][#1][c]{\linewidth}\vspace{0pt}%
		\lstset{style=codeinrow}%
	}{%
	\end{minipage}%
}

\newcommand{\PlotBox}[2][\FixedHeight]{%
	\begin{minipage}[c][#1][c]{\linewidth}\vspace{0pt}\centering
		\includegraphics[width=\linewidth,height=\dimexpr#1-0.6em\relax,keepaspectratio]{#2}
	\end{minipage}%
}

\newcommand{\NumCell}[2][\FixedHeight]{%
	\begin{minipage}[c][#1][c]{\linewidth}\vspace{0pt}\centering #2\end{minipage}%
}
\newcommand{\TextCell}[2][\FixedHeight]{%
	\begin{minipage}[c][#1][c]{\linewidth}\vspace{0pt}#2\end{minipage}%
}

\newcolumntype{P}[1]{>{\raggedright\arraybackslash}m{#1}}
\newcolumntype{C}[1]{>{\centering\arraybackslash}m{#1}}

\def\BibTeX{{\rm B\kern-.05em{\sc i\kern-.025em b}\kern-.08em
    T\kern-.1667em\lower.7ex\hbox{E}\kern-.125emX}}
\begin{document}

\title{A Multimodal Conversational Agent for Tabular Data Analysis}

\author{\IEEEauthorblockN{Mohammad Nour Al Awad}
	\IEEEauthorblockA{
		\textit{ITMO University}\\
		Saint Petersburg, Russia \\
		MohammadNourAlAwad@itmo.ru}
	\and
	\IEEEauthorblockN{Sergey Ivanov}
	\IEEEauthorblockA{
		\textit{ITMO University}\\
		Saint Petersburg, Russia \\
		svivanov@itmo.ru}
	\and
	\IEEEauthorblockN{Olga Tikhonova}
	\IEEEauthorblockA{
		\textit{ITMO University}\\
		Saint Petersburg, Russia \\
		tikhonova\_ob@itmo.ru}
	\and
	\IEEEauthorblockN{Ivan Khodnenko}
	\IEEEauthorblockA{
		\textit{ITMO University}\\
		Saint Petersburg, Russia \\
		ivan.khodnenko@itmo.ru}
}

\maketitle
\begingroup
\renewcommand\thefootnote{}\footnote{
	© 2025 IEEE. Personal use of this material is permitted.
	Permission from IEEE must be obtained for all other uses.
}\addtocounter{footnote}{-1}
\endgroup

\begin{abstract}
	Large language models (LLMs) can reshape information processing by handling data analysis, visualization, and interpretation in an interactive, context-aware dialogue with users, including voice interaction, while maintaining high performance. In this article, we present Talk2Data, a multimodal LLM-driven conversational agent for intuitive data exploration. The system lets users query datasets with voice or text instructions and receive answers as plots, tables, statistics, or spoken explanations. Built on LLMs, the suggested design combines OpenAI Whisper automatic speech recognition (ASR) system, Qwen-coder code generation LLM/model, custom sandboxed execution tools, and Coqui library for text-to-speech (TTS) within an agentic orchestration loop. Unlike text-only analysis tools, it adapts responses across modalities and supports multi-turn dialogues grounded in dataset context. In an evaluation of 48 tasks on three datasets, our prototype achieved 95.8\% accuracy with model-only generation time under 1.7 seconds (excluding ASR and execution time). A comparison across five LLM sizes (1.5B–32B) revealed accuracy–latency–cost trade-offs, with a 7B model providing the best balance for interactive use. By routing between conversation with user and code execution, constrained to a transparent sandbox, with simultaneously grounding prompts in schema-level context, the Talk2Data agent reliably retrieves actionable insights from tables while making computations verifiable. In the article, except for the Talk2Data agent itself, we discuss implications for human–data interaction, trust in LLM-driven analytics, and future extensions toward large-scale multimodal assistants.
\end{abstract}


\section{Introduction}
Interacting with data often requires programming skills or statistical expertise, creating barriers for managers, analysts, and other non-technical users~\cite{Zheng2024MultimodalTableUnderstanding,Sundar2023cTBLS}. Natural language interfaces (NLIs) aim to improve this information seeking process by allowing users to query data conversationally~\cite{Li2024NLITabular,CIS2022}. At the same time, voice interfaces are becoming increasingly common in daily life, yet existing voice assistants remain limited: they can answer factual questions or control devices, but they lack the analytical capabilities needed for meaningful data exploration.

LLMs now provide a powerful foundation for code generation and complex reasoning~\cite{Chen2021Codex,Roziere2023CodeLlama,Hui2024Qwen25}. Systems such as OpenAI’s Code Interpreter~\cite{OpenAI2023CodeInterpreter} demonstrate this potential, but they typically support only text-based input/output. Another problem is that features like multimodal responses and transparent execution of generated code (capabilities that are increasingly important for reliable human–data interaction in the information seeking process) are less present in current voice assistant designs.

We present Talk2Data, a multimodal conversational agent that enables users to effectively seek, retrieve, and analyze data stored in tabular datasets through either voice or text instructions, and to receive answers and insights as plots, tables, or spoken explanations. For example, users can ask “What’s the average delay for United flights?” or “Plot a histogram of age,” and the agent dynamically determines whether to generate Python code or provide a direct natural language response. Code is executed in a secure sandbox, results are narrated through text-to-speech (TTS), and multi-turn dialogue is supported via conversational memory. This design allows to adapt its responses to user intent, offering both analytical depth and accessibility.

At the core of this agent is a orchestration module that reasons about each query’s intent and selects the appropriate response path. Dataset metadata and conversation history are injected into structured prompts, enabling grounded, context-aware behavior. By integrating these components, this design demonstrates how LLMs can support natural, multimodal workflows for data analysis.

\textbf{Our contributions are as follows:}
\begin{itemize}
	\item \textbf{Multimodal information-seeking agent.} An end-to-end system that unifies voice/text input with visual, tabular, and spoken outputs for exploratory analysis, supporting multi-turn, back-and-forth dialogue for information seeking over tabular corpora.
	\item \textbf{Orchestration with transparent code execution.} A router that adaptively selects narration vs.\ code generation within one dialog using grounded prompts, paired with a secure sandbox that exposes code for provenance.
	\item \textbf{Empirical study across models and tasks.} A 48-task benchmark on three public datasets evaluating five \texttt{Qwen2.5-Coder} sizes to show how model scale affects information-seeking quality. The best mid-size configuration attains \textbf{95.8\%} accuracy with \emph{model-only} latency of \textbf{1.15--1.64\,s} (excluding ASR and execution time), and we analyze failure modes due to routing vs.\ runtime guardrails.
\end{itemize}

The remainder of this paper is organized as follows. Section~\ref{sec:related_work} reviews related research. Section~\ref{sec:system_arch} details the system architecture. Section~\ref{sec:experiments_benchmarks} presents experiments and benchmarks. Section~\ref{sec:discussion} discusses implications, Section~\ref{sec:limitations} outlines limitations. Section~\ref{sec:ethics} discusses ethical and responsible-AI considerations, and Section~\ref{sec:conclusion} concludes with future directions.

\section{Related Work}
\label{sec:related_work}

We review prior research in three areas most relevant to our work: (i) NLIs for data analytics and information retrieval, (ii) LLM-based code generation for data science, and (iii) voice-enabled data assistants.

\textbf{Text-Based NLIs.} Commercial tools such as Tableau Ask Data~\cite{Tableau2025AskData}, Power BI Q\&A~\cite{Microsoft2025PowerBIQA}, AWS QuickSight Q~\cite{AWS2025QuickSightQ}, and Qlik Sense~\cite{Qlik2025InsightAdvisor} enable querying structured data in natural language~\cite{Zheng2024MultimodalTableUnderstanding,Sundar2023cTBLS,Li2024NLITabular}. These interfaces lower the barrier to entry for exploration, but remain limited to text-based input/output. Recent surveys document the resurgence of neural methods for conversational information retrieval, highlighting the growing importance of multi-turn dialogue in facilitating interaction~\cite{Gao2022CIRSurvey, Hambarde2023IRSurvey,IRBook2024,Zhang2024AIRetrieval}. More recent LLM-powered utilities such as Powerdrill AI and ChatCSV~\cite{Powerdrill2025CSVAssistant,ChatCSV2025} automate chart generation, but still do not support spoken interaction or adaptive response modalities.

\textbf{LLM-Based Code Generation.} Advances in program synthesis with LLMs—including OpenAI Codex~\cite{Chen2021Codex}, Code Llama~\cite{Roziere2023CodeLlama}, and Qwen-2.5-Coder~\cite{Hui2024Qwen25}—have demonstrated strong performance in generating analysis pipelines and data visualizations from text instructions~\cite{Nascimento2024LLM4DS} which largely aids the process of retrieving insights using the generated code. However, these models are typically deployed in static, text-only settings. They lack conversational memory, execution tool, transparency of execution, and the ability to adapt outputs to user context or modality preferences. 

\textbf{Voice-Enabled Data Assistants.} Early prototypes built with LangChain, Streamlit, and gTTS~\cite{Ngonidzashe2023LangChainDemo,Kumar2023VoiceViz,GomezVazquez2024AutomaticGeneration,Zhao2023QTSumm} have demonstrated proof-of-concept voice interaction with tabular data. While promising, these systems remain limited: they provide only one-way voice interfaces, offer minimal multimodal feedback, and rely on heuristics rather than agentic decision-making about when to generate code versus when to respond conversationally.

\subsection{System Comparison}
We contrast representative systems along input/output modality, execution model, memory, and adaptivity.
Table~\ref{tab:comparison} summarizes key properties of existing systems. Unlike prior work, our design uniquely integrates (1) dual input and output modalities (voice and text), (2) LLM-powered code generation with safe and transparent sandboxed execution, (3) adaptive modality switching, i.e., flexibly choosing between charts, tables, or brief spoken/text responses within one dialogue, and (4) conversational memory for multi-turn analysis.

\begin{table*}[t]
	\centering
	\caption{Comparison of representative voice/data analytics systems.}
	\label{tab:comparison}
	\resizebox{\textwidth}{!}{%
		\begin{tabular}{lccccccc}
			\toprule
			\textbf{System} & \textbf{Input Modality} & \textbf{Execution} & \textbf{Visualization} & \textbf{Multimodal} & \textbf{Memory} & \textbf{Adaptive} & \textbf{License / Deploy} \\
			\midrule
			Tableau Ask Data & Text & — (no code) & Yes & No & Yes & No & Prop.; On\!-prem \\
			Power BI Q\&A & Text & — (no code) & Yes & No & Partial & No & Proprietary; SaaS \\
			ChatCSV & Text (CSV) & — (auto charts) & Yes & No & No & No & Proprietary; SaaS \\
			Powerdrill AI & Text & Auto analytics & Yes & No & Yes & No & Proprietary; SaaS \\
			LangChain Demos & Voice + Text & Prototype code gen & Basic & Partial & Yes & No & Open-source; On\!-prem \\
			Amazon QuickSight Q & Text & Auto analytics & Yes & No & Yes & No & Proprietary; SaaS \\
			Qlik Sense & Voice + Text & Auto suggestions & Yes & No & Yes & No & Proprietary; SaaS + On\!-prem \\
			\textbf{Talk2Data (Ours)} & Voice + Text & \textbf{Secure code execution} & \textbf{Yes} & \textbf{Voice \& Visual} & \textbf{Yes} & \textbf{Yes} & Open-source; On\!-prem \\
			\bottomrule
		\end{tabular}
	}
\end{table*}

We position our work within prior art by noting that earlier systems typically emphasize one or two axes of the problem: (i) text-only NLIs that translate simple questions into charts or aggregations~\cite{Tableau2025AskData,Microsoft2025PowerBIQA,AWS2025QuickSightQ,Qlik2025InsightAdvisor,Powerdrill2025CSVAssistant,ChatCSV2025}, (ii) LLM-based code synthesis for analysis and visualization~\cite{Chen2021Codex,Roziere2023CodeLlama,Hui2024Qwen25,Nascimento2024LLM4DS}, or (iii) voice front-ends that surface data with limited analytics depth~\cite{Ngonidzashe2023LangChainDemo,Kumar2023VoiceViz,GomezVazquez2024AutomaticGeneration,Zhao2023QTSumm}. Several recent demos combine two of these (e.g., voice \emph{plus} auto-charts, or text \emph{plus} code execution), but typically stop short of a unified conversation loop with memory, guarded execution, and modality-aware responses.

Another dimension of differentiation is deployment: most commercial NLIs such as Tableau Ask Data, Power BI Q\&A, or Amazon QuickSight Q are proprietary SaaS features with limited or no on-prem support. By contrast, our Talk2Data prototype follows an open-source, self-hostable trajectory, emphasizing transparency and feasibility for on-prem deployments.

The suggested design stays within this lineage and assembles known components—ASR, an LLM, and TTS—into a single workflow. The incremental aspects we explore are: (1) an explicit routing step (via a lightweight LangGraph node) that decides between code generation and direct narration using dataset-aware prompts; and (2) a restricted execution path that treats LLM code as a first-class artifact but runs it with guardrails (timeouts, whitelisting) and returns structured errors. Neither routing nor sandboxing is novel in isolation; our contribution is an empirical look at how these choices affect \emph{end-to-end} usability (accuracy, latency, and dialogue continuity) for tabular analysis, including when speech is the entry point.

Compared with text-only NLIs in commercial business intelligence (BI) tools~\cite{Tableau2025AskData,Microsoft2025PowerBIQA,AWS2025QuickSightQ,Qlik2025InsightAdvisor}, the suggested design emphasizes transparency and adaptability over enterprise-specific heuristics. Whereas BI platforms often rely on curated semantic layers, governed vocabularies, and policy enforcement, our system favors flexibility by exposing generated code and supporting free-form queries across arbitrary tabular datasets. This design trades some enterprise features (e.g., lineage tracking, synonym governance) for openness and interpretability—qualities important for exploratory analysis and educational use.

Relative to LLM “code interpreter” environments~\cite{OpenAI2023CodeInterpreter,Nascimento2024LLM4DS}, our scope extends by adding dual input/output modalities (voice and text, visuals and speech) and an explicit routing layer that determines when to generate code and when to respond conversationally. These choices enable multimodal, multi-turn interaction rather than single-shot code execution.

Voice-centric prototypes~\cite{Ngonidzashe2023LangChainDemo,Kumar2023VoiceViz,GomezVazquez2024AutomaticGeneration,Zhao2023QTSumm} demonstrate the feasibility of ASR–LLM–TTS pipelines but treat prompts as one-shot. By contrast, this design integrates a guarded execution environment with explicit error surfacing and conversational memory, supporting iterative refinement of queries while maintaining predictability and safety.

In that sense, this work demonstrates how combining speech recognition, LLM-based reasoning, guarded code execution, and text-to-speech within a single orchestration loop can yield a dependable multimodal assistant for tabular data. The suggested design offers an open-ended yet safe exploration of datasets through natural conversation, which forms a path toward more accessible and trustworthy multimodal data assistants.

\section{System Architecture}
\label{sec:system_arch}

Figure~\ref{fig:user_flow} presents the architecture and turn-level interaction loop. A user query enters a conversational workflow that initializes the agent state, selects an action, and returns a structured response. The pipeline is organized into four conceptual stages: preparing inputs and dataset context, deciding how to answer (code vs.\ narration), executing with safeguards, and delivering multimodal outputs while updating conversation state.

\subsection{Input and Context Assembly}
\label{sec:input_context}

\begin{figure*}[t]
	\centering
	\includegraphics[width=\linewidth,height=\textheight,keepaspectratio]{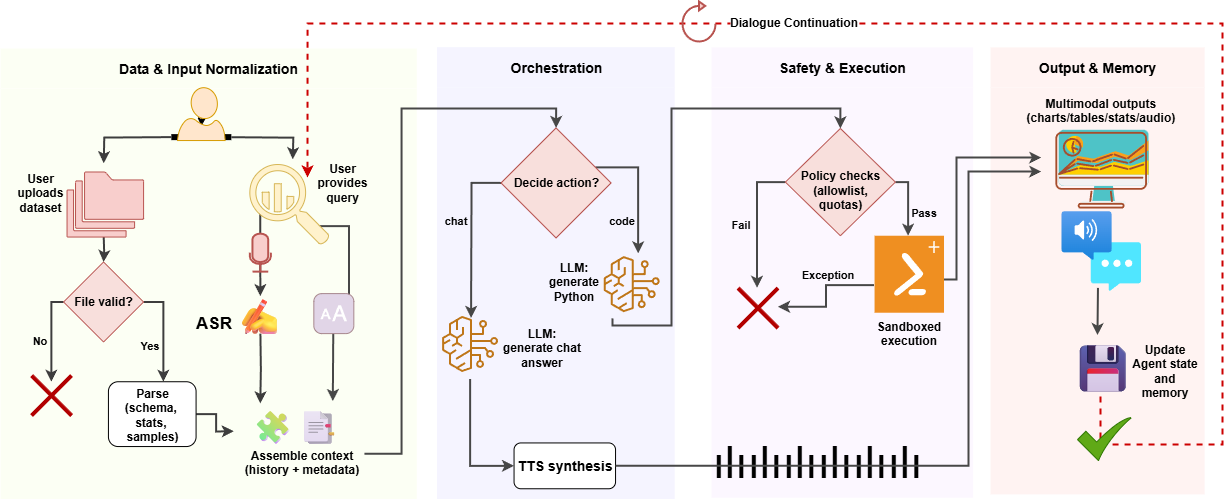}
	\caption{System and end-to-end user interaction in \emph{Talk2Data}. The router selects between chat and code paths; safety is enforced via sandboxed execution. Outputs may be visual, textual, or spoken, and the conversation state updates for subsequent turns.}
	\label{fig:user_flow}
\end{figure*}

This stage aligns two inputs—the dataset and the user’s utterance—into a shared, compact context that the rest of the pipeline can reliably consume (Fig.~\ref{fig:user_flow}). On the data side, we parse a lightweight schema and exemplars: column names and inferred types, a few sample rows, numeric ranges for quantitative fields, and representative values for categoricals. On the query side, both voice and text are normalized into a single text form (speech $\rightarrow$ ASR $\rightarrow$ text) using OpenAI's Whisper~\cite{Radford2022Whisper} for English speech; the ASR component is drop-in replaceable and can be extended to other languages by switching Whisper checkpoints or equivalent multilingual models.

The result is a \emph{context pack} that travels with the turn: it grounds routing (the decision prompt), constrains narration (the chat prompt), and anchors code synthesis (the code prompt) to what actually exists in the table. Ambiguities are reduced by aligning user words to schema tokens (e.g., resolving “GPA four” to \texttt{GPA4} when present) and by exposing recent conversation state so follow-ups (“now color by gender”) inherit prior choices without restating them.

Two invariants make this stage effective: (i) \emph{faithfulness}—only facts derived from the uploaded data (schema, ranges, exemplars) are injected; (ii) \emph{compactness}—the context is trimmed to remain stable across turns (e.g., small samples and capped categorical exemplars) so later stages see a consistent, prompt-friendly view. When the dataset cannot be parsed or violates basic expectations, the pipeline fails fast with a user-facing message rather than proceeding with an ungrounded context. This keeps subsequent decisions predictable and makes downstream successes attributable to a single, explicit source of grounding.

\subsection{Decision and Orchestration}
\label{sec:decision_orchestration}

At the core of the loop is a single decision point (Fig.~\ref{fig:user_flow}, \emph{decide\_action}) that classifies each turn as either \emph{code generation} or \emph{chat response}. Concentrating modality choice in one place keeps the rest of the pipeline simple: only one branch executes per turn, and every downstream component can assume a clear contract about what arrives next (either narration to render or code to check and run).

Orchestration is lightweight; it passes the same grounded context forward, selects the branch, and records the choice in the conversation state. This achieves three properties we rely on throughout the paper:
(i) \emph{Adaptivity} — the agent aligns response mode with user intent and dataset characteristics;
(ii) \emph{Predictability} — heavy operations (policy checks, execution) only occur when the decision warrants them;
(iii) \emph{Auditability} — each turn carries an explicit decision and rationale, so failures can be traced to a visible fork rather than hidden heuristics.

By separating \emph{decision} (what to do) from \emph{realization} (how it is done), the architecture admits evolution without churn: improving examples or style in the decision layer changes behavior immediately, while the execution and rendering layers remain stable. This division of concerns is what allows the whole design to be both adaptive in interaction and steady in operation.

\subsection{Prompt Design and Architectural Role}
\label{sec:prompt_design}

Prompting is the policy layer of our architecture: it encodes how the agent interprets intent, how it speaks, and how it produces computable artifacts—without hard-coding logic. The prompts operate at the three decision points highlighted in Figure~\ref{fig:user_flow}: (i) selecting the response mode, (ii) producing narration suited for TTS, and (iii) emitting code that downstream components can execute and capture reliably. Each prompt consumes the same structured context (dataset metadata and dialogue history) so that routing, language, and code are grounded in the same view of the data.

\paragraph*{Decision prompt}
This prompt helps the LLM to classify the user's request into \texttt{code\_generation} or \texttt{chat\_response}, returning the decision as minimal JSON. It is seeded with short, contrastive few-shot examples (``distribution'' $\rightarrow$ code; ``what columns'' $\rightarrow$ chat) to bias toward the most useful modality while still allowing clarification. Architecturally, this concentrates adaptivity into a single, auditable step: the rest of the pipeline remains simple because the mode is fixed before execution. When the request is ambiguous, the prompt favors a low-risk chat response and invites follow-up.

\paragraph*{Chat response prompt}
Here the policy is stylistic and accessibility-driven: produce brief, speakable text with no formatting, suitable for TTS, and grounded in the provided metadata/history. The prompt explicitly allows the assistant to ask clarifying questions and to propose switching to computation when the user’s intent is analytical. This preserves continuity (the conversation advances even when the dataset or request is unclear) and keeps narration aligned with what the code path would compute later.

\paragraph*{Code generation prompt}
This prompt constrains outputs to executable, comment-free Python that assumes a preloaded \texttt{df}. Two invariants matter architecturally: (1) \emph{grounding}—the code should reference only columns/types present in metadata; (2) \emph{capturability}—outputs end with an expression (e.g., a figure or dataframe variable) rather than prints. These constraints make execution predictable and simplify capture/rendering, while the few-shot patterns (simple histogram; head of a table) keep generations within the capabilities of the guarded runtime.

Embedding these policies in prompts yields a flexible control surface: we can adjust examples, brevity, or safety cues without modifying the orchestration logic. In practice, this reduced misclassifications and made failures legible—when pre-checks block an operation, the prompts already prime the assistant to explain the limitation and suggest a safer alternative.

\subsection{Code Generation and Secure Execution}

When the decision module selects the computational path, the agents constructs the code generation prompt. This prompt instructs the LLM to produce concise, expression-oriented Python code. The generated snippets typically leverage standard data analysis libraries such as \texttt{pandas}, \texttt{matplotlib}, \texttt{plotly}, or \texttt{seaborn}. By encouraging expression-based outputs rather than long scripts, the agent ensures that results are captured cleanly (e.g., figures, summary tables) and remain interpretable for users.

Generated code is passed to a sandboxed runtime that enforces strict constraints. Only whitelisted libraries are exposed, with file system, operating system, and network access disabled. Execution is bounded by timeouts and resource quotas to avoid denial-of-service or runaway computations. The pipeline captures outputs in structured form: numerical results are returned as JSON, and figures are converted to Base64 images that can be rendered in the user interface. This design also facilitates error handling: exceptions are intercepted, logged, and summarized into user-friendly explanations.

Allowing an LLM to generate and execute code introduces security and reliability risks. Without safeguards, generated code could accidentally overwrite files, exhaust memory, or leak private data. This sandbox design prioritizes safety and transparency by restricting system calls, limiting resource usage, and surfacing errors explicitly. This is a trade-off: some advanced operations (e.g., dynamic imports, external APIs) are not supported, which occasionally leads to task failures (see Section~\ref{sec:experiments_benchmarks}). However, we argue that constrained but safe execution is preferable to unrestricted but opaque pipelines, particularly for non-technical users who may not detect silent errors.

This secure execution pipeline provides several advantages to users. First, it builds trust: users can see and verify the code that produced their results, rather than relying on a “black box.” Second, it supports transparency: errors and limitations are explained instead of hidden, encouraging realistic expectations. Third, it maintains reproducibility: since outputs are generated from explicit code, they can be re-run or adapted manually by analysts.

This design treats code generation not as a hidden mechanism, but as a transparent, auditable process. This aligns with our broader design principle of \emph{trustworthy AI}, ensuring that powerful LLM-driven analysis remains safe, interpretable, and user-centered.

\subsection{Narrative Response and Dialogue Management}

When the agent chooses the \texttt{narrative response} path, it advances the conversation without running code: it interprets the user’s request against the current dataset context and recent turns, then produces a brief, speakable explanation. The aim is to keep momentum—clarify what a field means, summarize what a prior chart already showed, or resolve ambiguity with the smallest possible clarification—while remaining grounded in the same context that drives the analytic path.

Two short examples illustrate the role of this path:
\begin{quote}\small
	\textbf{User:} “What does \textit{carrier} mean here?”\\
	\textbf{Agent:} “It’s the airline operating the flight, like UA or DL, recorded as a short code.”
\end{quote}

\begin{quote}\small
	\textbf{User:} “How did you compute the on-time rate by carrier?”\\
	\textbf{Agent:} “I filtered out canceled flights, treated arrival delay $\leq 15$ minutes as on-time, grouped rows by carrier, and divided the on-time count by the total flights for each carrier.”
\end{quote}

Such answers reflect interpretive reasoning over metadata and align with human expectations. In this way, the agent not only delivers results but also explains them in terms accessible to non-specialists.

Providing natural explanations addresses a common limitation of analytic systems: non-technical users may not understand why a result looks the way it does. By embedding interpretive narratives into the workflow, this design increases trust, usability, and inclusivity. This functionality also enables accessibility scenarios (e.g., blind users receiving spoken feedback), underscoring the value of a multimodal conversational design.

In our prototype, narrative turns are rendered to audio via Coqui TTS (\texttt{tts\_models/en/vctk/vits}, speaker \texttt{p260})~\cite{CoquiTTS2023, Kim2021VITS}

\subsection{Multi-turn Interactions}

To support multi-turn conversations, the agent maintains a lightweight dialogue memory that includes active dataset metadata and a running history of prior queries and outputs (code or explanations). This context is injected into each new LLM prompt, allowing the agent to generate coherent, contextually aware responses. This design allows users to interact iteratively without restating full queries. For example:

\vspace{4pt}
\begin{quote}\small
	\textit{“Plot math vs.\ reading score.”} $\rightarrow$ [scatter plot] \\
	\textit{“Now color by gender.”} $\rightarrow$ [updated plot with hue]
\end{quote}
\vspace{4pt}

Here, the second query is resolved only by referencing the prior visualization and metadata. Such incremental refinement is characteristic of real-world analysis workflows and is not supported by one-shot NLIs.

We note that maintaining conversational memory introduces challenges: context windows are finite, and long interaction histories can increase latency or induce hallucinations. To mitigate this, we keep only last N interactions in the history. In future work, we plan to explore semantic memory compression, episodic summaries, and retrieval-augmented prompting to handle longer and more complex sessions.

Memory transforms the design from a query engine into a conversational partner. Users can build on prior steps naturally, explore alternative perspectives, and carry forward insights across turns. This supports more exploratory and human-like interaction with data.

Together, these capabilities—natural explanations, conversational memory, and multimodal output—differentiate this design from text-only NLIs and prototype voice demos, supporting the idea of a general-purpose conversational agent for data analysis.

Table~\ref{tab:refine_insurance} illustrates a three-turn refinement on the insurance dataset: the agent plots charges vs. BMI, colors by smoker status, then adds a regression line.

\begin{table*}[t]
	\centering
	\caption{Stepwise refinement scenario on an \textit{insurance} dataset: user command, generated code, and resulting plot.}
	\label{tab:refine_insurance}
	\setlength{\tabcolsep}{6pt}
	\renewcommand{\arraystretch}{1.08}
	
	\begin{tabular*}{\textwidth}{@{\extracolsep{\fill}}
			C{0.04\textwidth} P{0.1\textwidth} P{0.5\textwidth} C{0.3\textwidth}}
		\toprule
		\textbf{Step} & \textbf{User Command} & \textbf{Generated Code (snippet)} & \textbf{Resulting Plot} \\
		\midrule
		
		\NumCell{\textbf{1}} &
		\TextCell{Plot charges vs BMI} &
		\begin{FixedCodeBox}
			\begin{lstlisting}
				
plt.scatter(df['bmi'], df['charges'])
plt.xlabel('BMI')
plt.ylabel('Charges')
plt.title('Charges vs BMI')
plt.show()
			\end{lstlisting}
		\end{FixedCodeBox}
		&
		\PlotBox{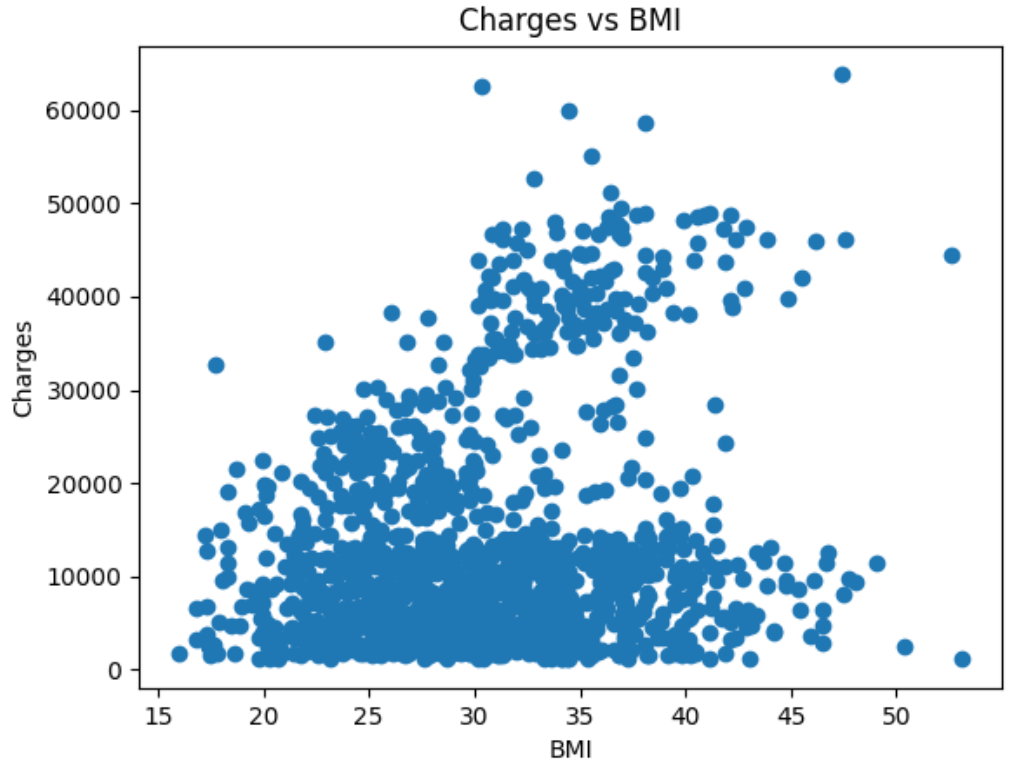}
		\\
		
		\NumCell{\textbf{2}} &
		\TextCell{now color by smoker status} &
		\begin{FixedCodeBox}
			\begin{lstlisting}
				
plt.scatter(df['bmi'], df['charges'], c=df['smoker'].map({'yes': 'red', 'no': 'blue'}))
plt.xlabel('BMI')
plt.ylabel('Charges')
plt.title('Charges vs BMI by Smoker Status')
plt.legend(['Smoker', 'Non-Smoker'])
plt.show()
			\end{lstlisting}
		\end{FixedCodeBox}
		&
		\PlotBox{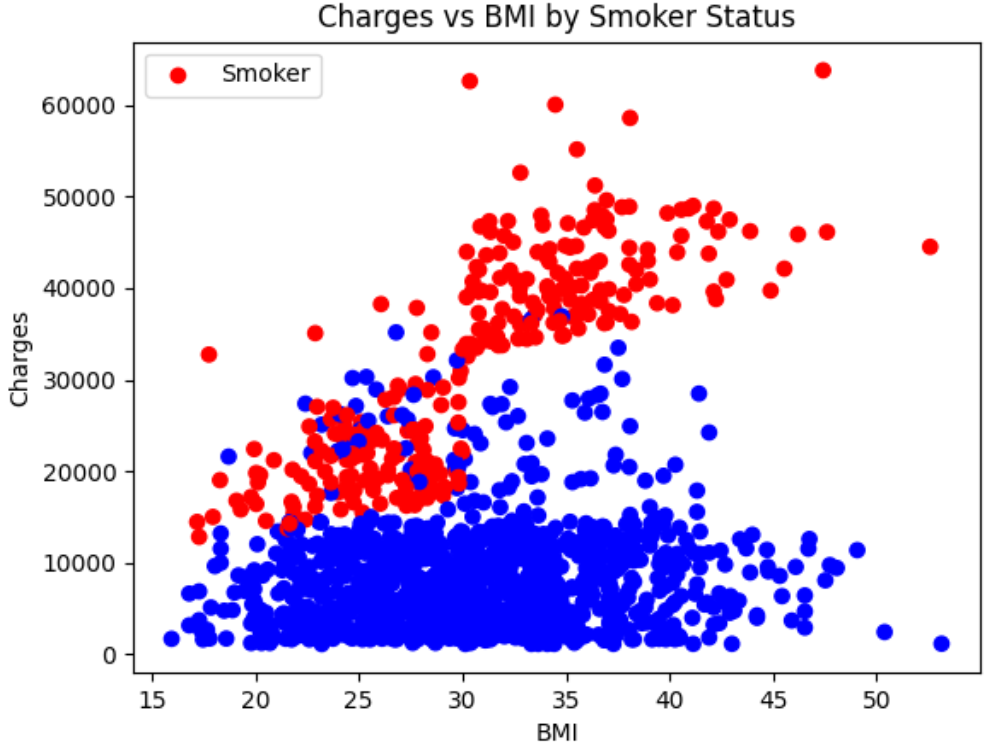}
		\\
		
		\NumCell{\textbf{3}} &
		\TextCell{now add a regression line} &
		\begin{FixedCodeBox}
			\begin{lstlisting}
plt.scatter(df['bmi'], df['charges'], c=df['smoker'].map({'yes': 'red', 'no': 'blue'}))
plt.xlabel('BMI')
plt.ylabel('Charges')
plt.title('Charges vs BMI by Smoker Status')
x = df['bmi']
y = df['charges']
slope, intercept = np.polyfit(x, y, 1)
plt.plot(x, slope*x + intercept, color='green')
plt.legend(['Smoker', 'Non-Smoker', 'Regression Line'])
plt.show()\end{lstlisting}
		\end{FixedCodeBox}
		&
		\PlotBox{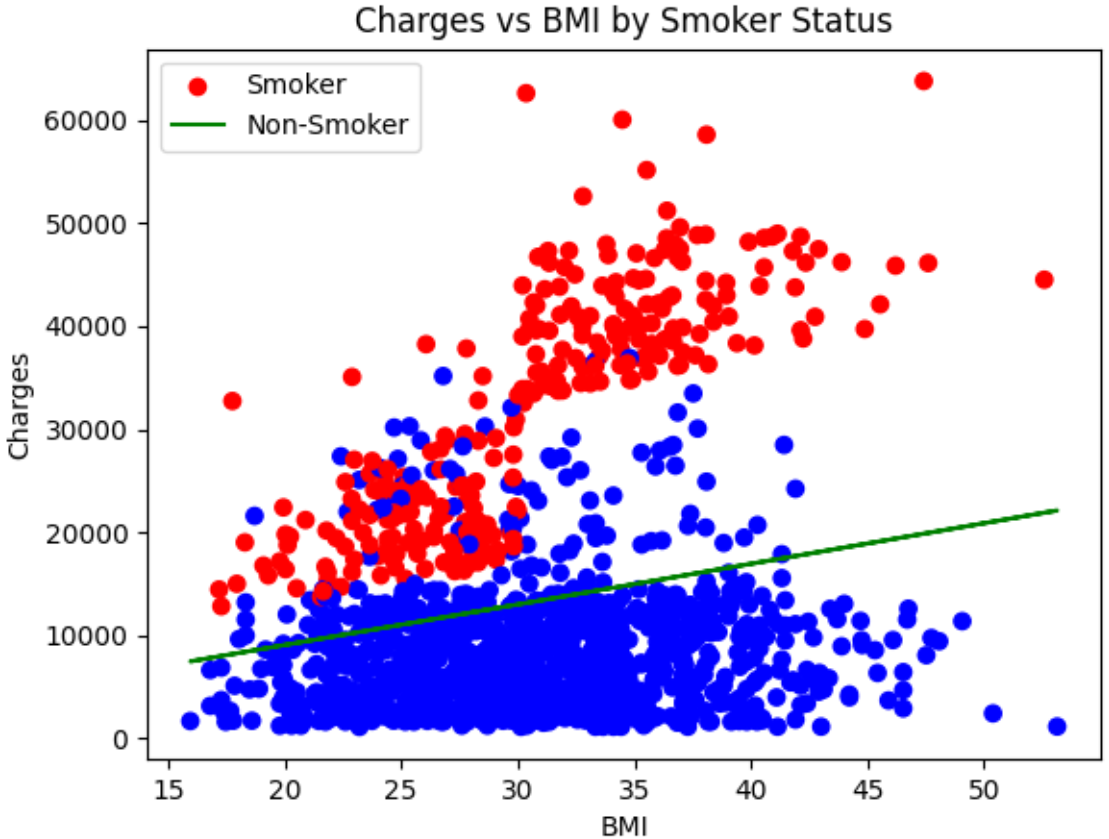}
		\\
		
		\bottomrule
	\end{tabular*}
\end{table*}

\section{Experiments \& Benchmarks}
\label{sec:experiments_benchmarks}

We evaluated the developer Talk2Data prototype on an NVIDIA RTX 6000 Ada (48\,GB VRAM) server, assessing both (1) \textbf{task performance} (accuracy on varied query types), and (2) \textbf{latency} (end-to-end response time). We also report qualitative observations of multi-turn behavior and decision-making. 

\subsection{Benchmark Design}

We tested on three public datasets (Table~\ref{tab:datasets}) spanning different sizes and complexities: the Otto Group product dataset~\cite{Otto2019Dataset}, US Flights 2008~\cite{Dongre2015USFlightsData}, and a student performance dataset~\cite{SPSCIENTIST2016StudentsPerformance}. Each dataset was evaluated with 16 natural language tasks, totaling 48.

\begin{table}[t]
	\centering
	\caption{Datasets used in evaluation.}
	\label{tab:datasets}
	\begin{tabular}{lrrl}
		\toprule
		\textbf{Dataset} & \textbf{Cols} & \textbf{Rows} & \textbf{Size} \\
		\midrule
		Otto Products     & 94 & 144{,}000     & $\sim$27\,MB \\
		Student Scores    & 8  & 1{,}000        & 0.07\,MB     \\
		US Flights 2008   & 29 & $\sim$7M       & 673\,MB      \\
		\bottomrule
	\end{tabular}
\end{table}

Tasks fell into three categories: 26 visualization requests (e.g., scatter plots), 13 analytical/statistical queries (e.g., finding max values), and 9 explanatory/narrative prompts (Table~\ref{tab:tasks}).

\begin{table}[t]
	\centering
	\caption{Evaluation query categories with examples.}
	\label{tab:tasks}
	\begin{tabular}{p{2.5cm} p{4.5cm}}
		\toprule
		\textbf{Category} & \textbf{Example Query} \\
		\midrule
		26 Visualization & "Create a scatter plot of departure vs.\ arrival delay with a trend line." \\
		13 Analytical    & "Find the product ID with the maximum value in feature 3." \\
		9 Narrative      & "Summarize the relationship between departure and arrival delays." \\
		\bottomrule
	\end{tabular}
\end{table}

Each task was evaluated manually. A query was marked correct if the output chart or response was accurate and aligned with expectations (based on manual ground truth calculations). Visualization tasks were also checked for proper axis labeling and scaling.

\paragraph*{Model variants and protocol.}
We evaluated five \texttt{Qwen2.5\text{-}Coder} instruction models (1.5B, 3B, 7B, 14B, and 32B-AWQ). Prompts, routing logic, datasets, and the guarded runtime were held constant across runs so that differences reflect the model’s contribution rather than orchestration or policy changes. Each task was executed once per model. We report: (i) \emph{Code} and \emph{Chat} correctness; (ii) \emph{Misclassification} when the router chose the wrong path; and (iii) \emph{Execution errors} when the sandbox blocked or the code failed at runtime.

\subsection{Accuracy and Model Trade-offs}

Table~\ref{tab:model_comparison} summarizes accuracy across models. Performance increases sharply from 1.5B$\rightarrow$3B$\rightarrow$7B, then shows diminishing returns: 7B and 32B-AWQ both reach 95.8\% overall, while 14B edges higher to 97.9\%.

\begin{table}[t]
	\centering
	\caption{Performance of \texttt{Qwen2.5-Coder-*B-Instruct} across 48 tasks (39 code + 9 chat).}
	\label{tab:model_comparison}
	\resizebox{\columnwidth}{!}{%
		\begin{tabular}{lcccccc}
			\toprule
			Model & Code & Chat & Misclassification & Exec errors & Overall \\
			\midrule
			1.5B & 21/39 (54\%) & 9/9 & 17 & 1 & 30/48 (62.5\%) \\
			3B   & 36/39 (92\%) & 6/9 & 3 & 3 & 42/48 (87.5\%) \\
			7B   & 37/39 (95\%) & 9/9 & 1 & 1 & 46/48 (95.8\%) \\
			32B-AWQ & 37/39 (95\%) & 9/9 & 0 & 2 & 46/48 (95.8\%) \\
			14B  & 38/39 (97\%) & 9/9 & 0 & 1 & 47/48 (97.9\%) \\
			\bottomrule
		\end{tabular}%
	}
\end{table}

We notice that smaller models behaved conservatively. The 1.5B variant often sought clarification or proposed how to proceed instead of producing code—counted as routing mistakes (17). When it did generate code, artifacts met minimal criteria but were simplified (e.g., default binning, missing groupings/labels), yielding 21/39 on code vs.\ 9/9 on chat; many turns ended in narration without an artifact.

At 3B, code correctness rose to 36/39, but the generated programs were generally simplistic (e.g. default setting for plotting), and chat dropped to 6/9: several narrative turns were terse or weakly grounded, reflecting a bias toward computation when a short explanation would have sufficed.

The 7B model had the most preferable results: one routing mistake, 37/39 code and 9/9 chat, with residual errors attributable to guardrails or environment (e.g., blocked imports) rather than misunderstanding.

Larger models offered better returns under our sandbox. The 14B model reached 38/39 code for the best overall score; 32B-AWQ matched 7B overall but showed slightly more execution errors (2), both rose from ambitious code blocked by the sandbox, i.e. attempting functionality beyond its restrictions.

In practice, the 7B model balances reliability and responsiveness at a substantially lower cost than larger variants. Specifically, the two failures of the 7B model were as follows:

\begin{itemize}
	\item \textbf{PCA failure:} A PCA scatter plot request failed because scikit-learn’s \texttt{PCA} import was blocked by the sandbox (no dynamic imports), illustrating that execution restrictions were correctly enforced.
	\item \textbf{Routing mistake:} A clear plotting request was classified as a \texttt{narrative response}; the assistant asked for clarification instead of generating code, so no artifact was produced and the task was marked incorrect.
\end{itemize}

These two failures reflect distinct modes: one intentional policy block by the runtime and one decision error at the routing step. Other visualizations—including histograms, boxplots, and correlation plots—were generated correctly. Analytical results (e.g., means, counts) matched expected values, and explanatory responses were grounded in the dataset metadata.

\subsection{Latency Analysis}

We measured per-query \emph{model-only time} across datasets. Here, \emph{model-only time} is defined as decision + (code generation \emph{or} chat response) + TTS, and \textbf{excludes} both automatic speech recognition (ASR) and code execution.

\begin{table*}[tb]
	\centering\small
	\caption{Average latency per query (seconds) across stages and datasets. The \textbf{Model-only time} column equals Decision + (Code Gen.\ or Chat Resp.) + TTS.}
	\label{tab:latency}
	\begin{tabular}{lcccccc}
		\toprule
		\textbf{Dataset} & \textbf{Decision} & \textbf{Code Gen.} & \textbf{Exec.} & \textbf{Chat Resp.} & \textbf{TTS} & \textbf{Model-only time (no ASR/exec.)} \\
		\midrule
		Products (medium) & 0.30 & 0.69 & 1.82 & 0.66 & 1.78 & 1.33 \\
		Students (small)  & 0.43 & 0.56 & 0.06 & 1.01 & 2.96 & 1.64 \\
		Flights (large)   & 0.32 & 0.57 & 3.82 & 0.45 & 1.13 & 1.15 \\
		\bottomrule
	\end{tabular}
\end{table*}

Decision time averaged \(0.30\text{--}0.43\)\,s, with code generation at \(0.56\text{--}0.69\)\,s. Sandbox execution (reported for context, not included in model-only time) scaled with dataset size—\(0.06\)\,s (Students), \(1.82\)\,s (Products), and \(3.82\)\,s (Flights). Chat responses took \(0.45\text{--}1.01\)\,s, and TTS added roughly \(1.1\text{--}3.0\)\,s per utterance. The resulting \emph{model-only time} averaged \(1.15\text{--}1.64\)\,s across datasets, comfortably within interactive bounds. Further gains are likely from streaming TTS and light pipelining (e.g., overlapping execution with rendering).

ASR is omitted because our benchmark sessions were text-only; code execution is excluded because it runs in a client-side sandbox and is highly machine-dependent. Thus, \emph{model-only time} should be interpreted as the latency to generate a response (code or narration) and synthesize speech, not the full end-to-end duration including execution.

Formally, the average \emph{model-only time} is
\[
T_{\text{model}} \;=\;
t_{\mathrm{dec}} \;+\;
\frac{ t_{\mathrm{code}} \cdot N_{\mathrm{code}} \;+\; (t_{\mathrm{chat}} + t_{\mathrm{tts}}) \cdot N_{\mathrm{chat}} }{ N_{\mathrm{total}} },
\]
where \(t_{\mathrm{dec}}\) is decision latency, \(t_{\mathrm{code}}\) is average code-generation time, \(t_{\mathrm{chat}}\) is average narrative generation time, \(t_{\mathrm{tts}}\) is average text-to-speech time, \(N_{\mathrm{code}}\) and \(N_{\mathrm{chat}}\) are the counts of code and chat turns, and \(N_{\mathrm{total}} = N_{\mathrm{code}} + N_{\mathrm{chat}}\).

We expect additional reductions in \(T_{\text{model}}\) from streaming TTS and stage pipelining (e.g., rendering while narration is being synthesized).

\section{Discussion} 
\label{sec:discussion}

Our evaluation surfaces where a conversational, multimodal analysis agent such as Talk2Data provides tangible advantages over text-only NLIs and one-shot code interpreters.

\subsection{Design objectives}

The architecture was shaped by several design goals that guided the technical choices:

\textbf{Accessibility.} By lowering entry barriers for non-technical users by supporting both text and voice queries, multimodal outputs, and conversational memory. This enables managers, analysts, and students to interact with data without writing code.

\textbf{Trust and safety.} Since the agent executes LLM-generated code, ensuring safety was a core requirement. We enforce this through sandboxed and transparent execution, and restrictions on file system and network access. These measures build user confidence and align with responsible AI principles.

\textbf{Adaptivity.} Unlike static dashboards or text-only assistants, this agent design adapts to user intent by switching modalities (charts, tables, speech). This adaptivity makes interaction more efficient and natural. An alternative axis of adaptivity, complementary to ours, is routing among search strategies for retrieval-augmented generation (RAG) pipelines to trade off quality and cost in multi-turn settings~\cite{Wang2024SRSA}.

\textbf{Extensibility.} The architecture was designed to be modular. Components such as the LLM, ASR, or TTS engine can be replaced with improved models. The agentic orchestration layer also allows easy integration of additional tools (e.g., retrieval modules, domain-specific analytics).

These principles emphasize that this design is a blueprint for building reliable, user-centered multimodal agents. They provide a foundation for future extensions, such as multilingual support, richer memory management, and domain-specific plugins.

\subsection{Implications for Human-Data Interaction}

The Talk2Data design shows that combining voice input with LLM-driven analysis makes data exploration more natural: users can ask open-ended questions and quickly obtain plots, tables, or brief spoken explanations. The high success rate (46/48) suggests current LLMs can handle diverse, average-difficulty queries on structured datasets.

Conversational memory supports exploratory analysis. Unlike one-shot NL2Viz systems, this design enables iterative probing—users ask a general question, then refine naturally—mirroring how analysts work and deepening understanding.

Users do not always know how to phrase a request or which format is ideal. The agent fills that gap: if a user asks, “What’s interesting in this dataset?”, it may produce a spoken summary and then offer a chart; follow-ups like “Plot that” continue seamlessly. Centralizing modality selection at the turn level simplifies interaction: the system commits to either narration or computation, avoiding partial, mixed responses. Code transparency makes results auditable; short, expression-final snippets connect outputs to causes and enable reuse.

\subsection{Broader usability}
\label{sec:use_cases}

While Talk2Data is presented as a research prototype, its design enables deployment in several practical scenarios where conversational and multimodal data access can add value. Below we outline representative use cases that illustrate the agent’s potential impact.

\begin{itemize}
	\item \textbf{On-the-go \& meetings:} Hands-free voice queries with spoken summaries and optional charts for quick check-ins.
	\item \textbf{Mobile / limited screens:} Concise answers and single-view visuals reduce navigation and typing.
	\item \textbf{Accessibility:} TTS explanations of distributions, trends, and outliers; screen-reader friendly flow.
	\item \textbf{Managerial support:} High-level KPIs, deltas, and ``what changed?'' prompts; easy export of figures.
	\item \textbf{Education \& training:} Explain--then--show interactions (e.g., ``What is a boxplot?'') with iterative follow-ups.
	\item \textbf{Collaborative sessions:} Shared visuals plus narration in real time; where memory enables fast refinements.
	\item \textbf{Analyst prototyping:} First-pass code/plots to copy, edit, and rerun in notebooks; reproducible snippets.
\end{itemize}

These scenarios illustrate how an agentic, multimodal interface for data analysis can benefit both technical and non-technical users. This also highlight the broader potential of LLMs in enabling accessible, trustworthy, and adaptive interaction with structured data. Importantly, the same design pattern can orchestrate other tools (e.g., SQL engines, geospatial queries, simulators, or document extractors), enabling end-to-end workflows that mix retrieval, transformation, visualization, and explanation.

\section{Limitations and Threats to Validity}
\label{sec:limitations}

This section consolidates the work’s limitations and threats to validity, clarifying scope, assumptions, and potential biases.

\textbf{Coverage and generalization.}
Our 48-task benchmark spans mostly clean tabular datasets. Performance on noisy, domain-specific, time-series, or hierarchical tables may differ. Extending to schema repair and retrieval-augmented prompting is future work.

\textbf{Speech recognition realism.}
Our evaluation excluded ASR by using text-only queries. In practice, real-world use introduces variability from accents, background noise, and device microphones, which may reduce accuracy or increase latency.

\textbf{Prompt and memory policies.}
Prompt templates and the lightweight dialogue memory (last-$N$ turns) may not transfer unchanged across model families or long sessions; context limits can induce omissions or hallucinations. Episodic summaries and policy re-tuning are likely needed for longer dialogues and other LLMs.

\textbf{Evaluation procedure and bias.}
Correctness was judged manually with ground-truth checks but without blinded raters; inter-rater reliability was not measured. Broader user studies (incl.\ non-technical participants and screen-reader users) are needed to assess usability, trust, and cognitive load.

\textbf{Scalability and cost.}
Serving ASR, TTS, and larger LLMs concurrently is resource-intensive. While 7B models offered a strong cost–quality point here, deployment may require quantization, speculative decoding, or tiered routing to smaller models.

\textbf{Security surface.}
Sandboxing blocks OS/network access and enforces time/CPU limits, but residual risks remain (e.g., resource exhaustion). Micro-VM isolation, stricter quotas, and policy checks for visual integrity can further reduce attack surface.

\textbf{Audio latency.}
TTS dominates short-turn latency in Talk2Data current design. Streaming/duplex audio (barge-in) and overlapping synthesis with rendering can reduce perceived delay substantially.

We plan to (i) support cross-dataset joins with retrieval and schema repair, (ii) add streaming TTS with interruption, and (iii) expose a plug-in API with micro-VM execution and resource quotas for even safer extensibility.

\section{Ethical and Responsible Considerations}
\label{sec:ethics}
Deploying multimodal agents such as our Talk2Data raises important ethical and responsible AI concerns. While our work focuses on technical feasibility, we highlight key dimensions that must be addressed when moving toward real-world deployment.

\textbf{Trust and transparency.} Users should understand how results are produced and when code is executed. This design promotes transparency by surfacing generated Python code. This helps users distinguish between correct outputs and system limitations.

\textbf{Data privacy.} Voice queries and dataset content may contain sensitive information. Deployments of such agent design should ensure strict data governance, including local execution or encryption of queries, and careful handling of conversational histories.

\textbf{Accessibility and inclusion.} A positive ethical dimension of this design is its ability to support users who are traditionally under-served by data tools, such as visually impaired individuals. However, accessibility should not come at the cost of accuracy or safety, and careful evaluation is needed to ensure inclusive benefit.

\textbf{Security risks.} Even with sandboxing, executing generated code introduces potential attack surfaces such as denial-of-service through resource exhaustion. Strengthening the sandbox with quotas, micro-VMs, or policy enforcement layers is critical to prevent misuse.

By foregrounding these considerations, we aim to emphasize that technical progress in multimodal agents must go with responsible design. We view Talk2Data not as a technical contribution but also as an opportunity to reflect on how agentic systems can be made trustworthy, transparent, and inclusive.

\section{Conclusion}
\label{sec:conclusion}
We presented \emph{Talk2Data}, a multimodal conversational agent for tabular analytics that combines ASR, LLM-based decision making and code generation, secure sandboxed execution, and TTS within a single orchestration loop. By centralizing the code-vs-chat decision in a lightweight router and grounding all prompts in a compact dataset context, the system delivers transparent, auditable analytics while remaining accessible to non-technical users.

Across 48 tasks on three public datasets, our prototype achieved 95.8\% accuracy with model-only latency within 1.15--1.64\,s, suggesting that careful assembly of existing components---paired with explicit guardrails and prompt policies---is sufficient for dependable, interactive analysis. Beyond raw accuracy, we find that conversational memory, and code transparency are key to user trust and iterative refinement. While not a new primitive, the design demonstrates a practical blueprint for building dependable, voice- and text-driven data assistants that prioritize safety, adaptability, and interpretability.

\subsection{Future Work and Challenges}
We see several directions to extend capability, robustness, and real-world utility:

\begin{itemize}

	\item \textbf{Broader data coverage.} Extend beyond clean tabular data to time series, relational joins, and schema mismatch via retrieval-and-rewrite (schema repair, type reconciliation) and light feature discovery.

	\item \textbf{Safety and isolation.} Harden execution with micro-VMs, stricter resource quotas, and a policy engine for API/library allowlists; add static analysis of generated code and visual-integrity checks on produced figures.

	\item \textbf{Richer memory and longer contexts.} Replace last-$N$ history with episodic summaries and retrieval-augmented prompting; explore semantic compression and tool-aware state to preserve salient decisions (filters, encodings) across long sessions.
	
	\item \textbf{Lower-perceived latency audio.} Add streaming ASR/TTS with barge-in and turn-taking cues; pipeline narration with rendering to reduce time-to-first-token and enable mid-utterance corrections.
	
	\item \textbf{Model efficiency.} Explore tiered routing (small model first, escalate on uncertainty), speculative decoding, quantization/LoRA variants, and knowledge distillation to maintain quality at lower cost.
	
	\item \textbf{User studies and evaluation.} Run controlled studies with non-technical users to measure task time, trust, and cognitive load; add blinded multi-rater correctness and scenario-driven stress tests.
	
\end{itemize}
	
Together, these steps move the system from a research prototype toward a deployable assistant that maintains transparency and safety while scaling to richer data, longer dialogues, and diverse users.

\section*{Artifact Availability}
The full implementation—including code, evaluation scripts, results, and a demo video—is available at:

\noindent\url{https://github.com/mohammad-nour-alawad/talk2data}

\section*{Acknowledgment}
This work supported by the Ministry of Economic Development of the Russian Federation (IGK 000000C313925P4C0002), agreement No139-15-2025-010

\bibliographystyle{IEEEtran}
\bibliography{custom}

\end{document}